\documentclass[11pt]{article}

\usepackage[final]{acl}

\usepackage{times}
\usepackage{latexsym}

\usepackage[T1]{fontenc}

\usepackage[utf8]{inputenc}

\usepackage{microtype}

\usepackage{inconsolata}

\usepackage{graphicx}

\usepackage{amsmath}
\usepackage{tikz}
\usetikzlibrary{positioning, shapes.geometric, arrows.meta, fit, backgrounds, calc}
\usepackage{amssymb}

\title{Overcoming the Impedance Mismatch: A Theoretical Roadmap for Fusing Foundation Models and Knowledge Graphs}

\author{Sahil Rajesh Dhayalkar \\
  Arizona State University \\
  \texttt{sdhayalk@asu.edu}\\
}

\begin{document}
\maketitle
\begin{abstract}
Modern artificial intelligence remains fundamentally divided between the continuous, probabilistic spaces of Foundation Models and the discrete, deterministic structures of Knowledge Graphs. While Retrieval-Augmented Generation (RAG) attempts to connect them by serializing graph data into text, we argue this lexical bridging is merely a superficial patch. In this paper, we formalize the underlying structural and geometric friction as the \textit{Impedance Mismatch}. By categorizing current neuro-symbolic integration strategies into a three-tiered hierarchy, we demonstrate that neither surface-level prompt injection nor continuous representation alignment can preserve the strict logical motifs required for reliable multi-hop reasoning. We define the specific mathematical limits, such as the Lexical Bottleneck and Topological Collapse, that show current architectures will eventually hallucinate or conflate semantic nodes. To achieve true semantic fusion, we propose a rigorous theoretical roadmap. We advocate for natively internalizing discrete symbolic structures through Structured Residual Streams, utilizing Vector Symbolic Architectures for latent sub-graph injection, and performing model updates via Orthogonal Subspace Editing. This actionable framework paves the way for models that seamlessly fuse the precision of symbolic logic with the expressivity of parametric memory.
\end{abstract}

\section{Introduction}
\label{sec:intro}

The architecture of modern artificial intelligence remains fundamentally divided by two distinct paradigms of knowledge representation. On one hand, the subsymbolic paradigm relies on the distributed, continuous representation spaces of Foundation Models, where transformer-based large language models~\citep{vaswani} represent vast amounts of probabilistic world knowledge during pre-training \citep{brown2020language, touvron2023llama, openai2024gpt4}. On the other hand, classical symbolic artificial intelligence utilizes discrete, structured formalisms like Knowledge Graphs to explicitly model declarative knowledge as rigid relational structures \citep{hogan2021knowledge, ji2021survey}. These symbolic frameworks inherently provide the explicit semantics, rigorous compositional structure, and strong mathematical guarantees regarding constraint satisfaction that standard neural architectures natively lack. Bridging this divide is recognized as the next step for Artificial General Intelligence (AGI) \citep{pan2024unifying, luo2025integrating}.

As foundational models are deployed in high-stakes, knowledge-intensive environments, the need to ground their parametric memory in reliable and up-to-date factual repositories has become critical \citep{li2026survey, ma2025large}. The prevailing industrial solution is Retrieval-Augmented Generation (RAG) \citep{lewis2020retrieval, guu2020realm, gao2023retrieval}. Current RAG methodologies typically attempt to bridge this gap by serializing knowledge graph subgraphs into natural language strings and injecting them directly into the context window of the model \citep{edge2024graphrag, xu2024retrieval}. However, we argue that this bridging strategy serves as a superficial patch rather than a mathematical structural solution. Treating the challenge of knowledge integration as mere text retrieval ignores the structural and geometric friction between discrete symbolic edges and continuous parameter spaces \citep{chen2025practices, wang2024large}. 

In this paper, we formalize this structural friction as the \textit{Impedance Mismatch} of neuro-symbolic knowledge integration. Borrowing a foundational concept from object-relational database theory, we define the impedance mismatch as the mathematical degradation that occurs when deterministic graph-structured knowledge bases are artificially mapped into probabilistic self-attention-driven latent spaces \citep{chen2025practices}. Foundational models perceive the world probabilistically through dense vector similarities, whereas databases and knowledge graphs require strict deterministic algorithmic manipulation. When large language models attempt to process standard knowledge graph structures, they struggle against their own continuous training priors \citep{wang2024large}. This conflict directly results in information loss driven by tokenization mismatches between LLM text encoders and discrete knowledge graph embeddings \citep{chen2025practices, pan2024unifying}. Furthermore, converting a rigid relational tuple into a linear sequence of tokens fails to preserve the relational geometry required for multi-hop logical reasoning, directly causing high non-retrieval rates, disconnected subgraphs, and hallucinations \citep{jiang2025reasoning, regraphrag2025, ma2025large, edge2024graphrag}.

To advance beyond the limitations of text-based retrieval frameworks and achieve true semantic fusion between foundational models and knowledge graphs, we attempt to provide a rigorous theoretical foundation. Our contributions are:
\begin{itemize}
    \item A Hierarchy of Integration Strategies: We propose a comprehensive hierarchy of integration strategies that categorizes current methodologies from lexical injection to architectural embeddings, highlighting the theoretical capacity limits of each paradigm \citep{ma2025large, wang2024large}.
    \item Identification of Core Bottlenecks: We define three bottlenecks preventing true neuro-symbolic fusion, specifically detailing the saturation limits of differentiable logic \citep{vankrieken2022analyzing}, the structural and geometric interference of continuous memory, and the fundamental asymmetry of symbol grounding \citep{harnad1990symbol, ji2021survey}.
    \item A Roadmap for the Knowledge Lifecycle: We chart a theoretical roadmap spanning the complete knowledge lifecycle of emergence, injection, and updating \citep{dhayalkar_roadmap}. We propose mechanisms like latent subgraph injection and orthogonal subspace editing to resolve the impedance mismatch directly within the transformer architecture, paving the way for verifiable compositional generalization \citep{pan2024unifying, luo2025integrating}.
\end{itemize}

Hence, we discuss that building knowledgeable foundation models requires moving beyond the assumption that continuous weights can seamlessly absorb discrete facts without explicit, mathematically grounded architectural mediation \citep{neurips2025synergizing, pan2024unifying}.

\section{The Anatomy of the Impedance Mismatch}
\label{sec:anatomy}

To understand why simple text-based retrieval fails to achieve true semantic fusion, we must establish the differences between symbolic graphs and continuous vector spaces. The core technical challenge of integration lies in reconciling the continuous, statistical nature of neural networks with the discrete, logical nature of symbolic systems \citep{garcez2022neural, ji2021survey}. We categorize this impedance mismatch across three structural dimensions: relational architecture, logical certainty, and memory editability.

\subsection{Formalizing the Impedance Mismatch}
\label{sec:formalization}

To ground the impedance mismatch, we must formalize the structural degradation that occurs when mapping discrete relational architectures into continuous latent spaces \citep{chen2025practices}. 

Let a Knowledge Graph be defined as a discrete topological space $\mathcal{K} = (\mathcal{V}, \mathcal{E})$, where $\mathcal{V}$ represents the set of entity vertices and $\mathcal{E}$ represents the set of relational edges. This space is equipped with a shortest-path metric $d_{\mathcal{K}}(v_i, v_j)$ that calculates the discrete logical distance between two entities $v_i, v_j \in \mathcal{V}$. Conversely, let the Foundation Model's latent space be a continuous metric space $\mathcal{M} \subseteq \mathbb{R}^h$, where $h$ denotes the dimensionality of the dense vectors, equipped with a geometric distance function $d_{\mathcal{M}}$. Any integration strategy requires a representation mapping function $f: \mathcal{V} \rightarrow \mathcal{M}$.

According to the principles of metric embedding theory, mapping an arbitrary discrete graph into a continuous vector space guarantees a strictly positive structural distortion. We formally define the Impedance Mismatch, denoted as $\mathcal{I}$, as the unavoidable mathematical lower bound of this distortion:
$$ \mathcal{I} = \inf_{f} \left( \sup_{u \neq v} \frac{d_{\mathcal{M}}(f(u), f(v))}{d_{\mathcal{K}}(u, v)} \times \right. $$
$$ \qquad \qquad \left. \sup_{u \neq v} \frac{d_{\mathcal{K}}(u, v)}{d_{\mathcal{M}}(f(u), f(v))} \right) $$
where $\inf_{f}$ denotes the infimum (greatest lower bound) over all possible mapping functions $f$, and $\sup_{u \neq v}$ denotes the supremum (least upper bound) over all distinct pairs of entities $u, v \in \mathcal{V}$. In a purely discrete, deterministic system, $\mathcal{I} = 1$, representing perfect structural isometry. However, for dense transformer representations, $\mathcal{I} \gg 1$. This formula shows that continuous spaces cannot faithfully preserve complex graph motifs, such as closed cycles and hierarchical trees, without warping the distances between nodes \citep{wang2024large}. 
Furthermore, this mismatch manifests as a compounding error during relational composition. In a discrete graph, navigating from a source entity $v_1$ to a target entity $v_3$ via sequential relations $r_1$ and $r_2$ is a deterministic algebraic composition, yielding an exact target node. In a foundation model, this multi-hop relation is approximated geometrically via sequential self-attention blocks. If $A^{(l)}$ represents the attention matrix at layer $l$, and $L$ represents the total number of attention layers, the continuous approximation introduces an error term $\epsilon$:
$$\epsilon = \left\lVert f(v_3) - \prod_{l=1}^{L} A^{(l)} f(v_1) \right\rVert$$
As the number of logical hops increases, the continuous approximation error $\epsilon$ compounds multiplicatively. This formalizes exactly why text-based retrieval frameworks fail at multi-hop logical reasoning \citep{jiang2025reasoning, regraphrag2025}: the continuous representation natively lacks the closed algebraic properties required to keep $\epsilon$ at zero.

\subsection{Structural versus Geometric Relations}

In a knowledge graph, knowledge is defined structurally. A relation between a subject entity $v_s$ and an object entity $v_o$ via a predicate $r$ is represented as an explicit, discrete edge $(v_s, r, v_o) \in \mathcal{E}$, where $\mathcal{E}$ is the set of all edges in the graph \citep{hogan2021knowledge}. Retrieving a fact or executing a multi-hop logical query relies on exact graph traversal. The expressive power of such representations depends heavily on the discrete structural motifs used to capture interactions.

Conversely, Foundation Models operate in continuous, high-dimensional vector spaces where internal states are represented by dense tensors \citep{brown2020language, touvron2023llama}. Relations are not explicit edges but are instead approximated geometrically through implicit affine transformations and attention-weighted sums. While a knowledge graph queries adjacency via an indicator function or boolean matrix multiplication, a transformer layer models a relation by computing a soft self-attention distribution \citep{vaswani}:
$$ \text{Attn}(Q, K, V) = \text{softmax} \left( \frac{QK^\top}{\sqrt{d_k}} \right) V $$
In this geometric space, the relational edge between two concepts is a dense similarity scalar in the attention matrix. This continuous perception struggles to preserve the strict structural constraints required for reliable, multi-step symbolic reasoning \citep{pan2024unifying, wang2024large}. When discrete graph architecture is forced into this continuous geometry, the crisp boundaries of symbolic motifs inevitably blur. This geometric blurring directly leads to hallucinated edges, invalid logical hops, and a degradation of verifiable inference \citep{jiang2025reasoning, luo2025integrating, edge2024graphrag}.

\subsection{Certainty versus Probability}

The second dimension of the mismatch concerns the truth representation of the encoded knowledge. Knowledge graphs are explicitly built on deterministic logic. An edge either exists or it does not, providing definitive, discrete representations of facts. This structural rigidity makes them suitable for precise querying and explainable, rule-based reasoning \citep{hogan2021knowledge, ji2021survey}.

However, foundational models are fundamentally probabilistic engines trained to minimize cross-entropy loss over token distributions to learn statistical regularities of language \citep{openai2024gpt4}. Their internal representation of a fact is inherently statistical and highly contextual. Real-world knowledge is thus modeled not as a binary truth but as a continuous probability density. Merging these two paradigms can cause a structural collapse \citep{pan2024unifying}. Either the definitive certainty of the knowledge graph must be relaxed into a probabilistic embedding, which mathematically destroys its logical guarantees, or the continuous parameter space of the foundational model must be artificially thresholded to accommodate discrete rules \citep{luo2025integrating, zhang2025enhancing}. Standard hybrid predictors often assume conditional independence between extracted symbols to bridge this gap. Unfortunately, this assumption limits their ability to model complex interactions and leads to overconfident, miscalibrated predictions \citep{wang2024large, jiang2025reasoning}.

\subsection{The Editability Conflict}

Another problem with this impedance mismatch is the difference in how the two systems update their information.
Knowledge graphs are highly dynamic and editable. Updating a fact or correcting an outdated relationship requires a straightforward $O(1)$ operation, executing the direct insertion or deletion of a discrete edge $(v_s, r, v_o)$ \citep{hogan2021knowledge}. 

Updating the parametric memory of a foundational model presents a very different theoretical challenge \citep{de2021editing, mitchell2022fast}. Knowledge in a transformer is heavily interconnected across multiple layers and attention heads via dense vector addition. Modifying a specific fact requires gradient descent or surgical weight perturbations, operations that are inherently unstable for lifelong editing \citep{meng2022locating, yao2023editing}. Recent studies in continuous knowledge editing reveal a significant performance decline in both knowledge update efficacy and retention as the number of sequential edits increases \citep{de2021editing, hase2024does}. Because the representations are continuous and overlapping, altering the parameters to update one fact often causes degraded interference with adjacent, structurally unrelated knowledge \citep{meng2022locating, yao2023editing, mitchell2022fast}. While novel techniques that disentangle and sparsify knowledge representations show promise in alleviating this decline, the fundamental editability conflict remains an unsolved barrier \citep{pan2024unifying, luo2025integrating}. The distributed nature of the embedding space inherently resists the localized, surgical updates that discrete knowledge graphs effortlessly support.

\section{A Hierarchy of Integration Strategies}
\label{sec:hierarchy}

To analyze neuro-symbolic research, we structure existing literature into a three-tiered maturity model. This hierarchy categorizes integration strategies based on how deeply the discrete knowledge graph penetrates the continuous architecture of the foundational model \citep{pan2024unifying, luo2025integrating, wang2024large}. As summarized in Table~\ref{tab:integration_limits}, we can then isolate and expose the specific theoretical limitations inherent to each paradigm.

\subsection{Level 1: Lexical and Prompt Injection (Surface-Level)}

The most common integration paradigm in industrial and academic settings operates entirely at the surface level. This is mostly realized through Knowledge Graph-Augmented Generation frameworks \citep{lewis2020retrieval, gao2023retrieval, xu2024retrieval, liu2025hyp}. In this approach, an external retriever isolates a structurally relevant subgraph, serializes the discrete triples into natural language text, and concatenates this verbalized payload directly into the context window of the foundational model \citep{lewis2020retrieval, chen2025gril}. Recent frameworks have attempted to optimize by retrieving hypothetical reasoning paths to improve evidence selection or by deploying adaptive multi-hop algorithms to reduce the overall token payload \citep{edge2024graphrag, liu2025hyp}.

Critique: While this methodology is accessible and deployable, lexical injection functions as a superficial patch. It inherently suffers from inference latency and remains bottlenecked by context window limitations. Surface-level integration is susceptible to knowledge conflicts, where the model's parametric memory overrides the retrieved context \citep{jiang2025reasoning, pan2024unifying}. When the verbalized graph information logically contradicts the pre-trained continuous weights of the foundation model, the architecture frequently discards the prompt in favor of its statistical prior \citep{mallen2023trust, xie2024adaptive, jiang2025reasoning}. Furthermore, serializing a complex multidimensional graph structure into a flat, linear token stream dismantles the structural motifs required for multi-hop logical deduction \citep{edge2024graphrag, chen2025practices}.

To formally demonstrate this limitation, we define the mathematical boundary of the Lexical Bottleneck. Let a knowledge subgraph $\mathcal{G} = (\mathcal{V}, \mathcal{E})$ possess an average branching factor $b$ and require a logical reasoning depth of $k$. Let $\mathcal{T}$ represent the token space of a foundational model with a maximum context window length $L$. Assuming a uniform average branching factor $b$, the number of distinct reasoning paths of length $k$ diverging from a source entity is $b^k$. The total number of elements required to fully represent this reasoning subgraph scales geometrically as $\mathcal{O}(b^k)$. 

If $c \ge 1$ is the minimum number of tokens required to serialize a single graph element, the minimum token length to represent the subgraph is $c \cdot \mathcal{O}(b^k)$. By the Pigeonhole Principle, if this required length exceeds the fixed capacity $L$, any deterministic serialization function must truncate information. In classical logic, removing a single premise from a multi-hop chain invalidates the entire deductive path. Consequently, as the reasoning depth $k$ scales, preserving the complete set of relational paths becomes mathematically impossible without unbounded information loss.

\subsection{Level 2: Representation Alignment (Embedding-Level)}

To bypass the tokenization bottlenecks of text verbalization, the second tier of integration attempts to align the representations of the knowledge graph and the foundational model within a shared latent mathematical space. Methodologies typically employ Graph Neural Networks or sophisticated translation-based embedding techniques to encode the relational architecture of the discrete graph into dense continuous vectors \citep{bordes2013translating, kipf2017semi, wang2024large, yasunaga2021qa}. These graph embeddings are then fused, concatenated, or aligned via multi-task contrastive learning objectives with the native text embeddings of the foundational model during an explicit forward pass or intermediate fine-tuning stage \citep{sat2025enhancing, luo2025integrating, zhang2025enhancing}.

Critique: Embedding-level alignment represents a significant step forward, yet it introduces a representational gap \citep{pan2024unifying}. Forcing a strict discrete graph into a continuous text embedding space necessitates a mathematical projection that degrades the strict relational properties of the original symbolic graph \citep{sat2025enhancing, chen2025practices}. In this paradigm, the continuous vector space acts as a lossy compression algorithm for discrete logic. The system permanently loses the precise relational boundaries inherent to discrete symbols. Hence, while the foundational model gains broad domain awareness, it remains incapable of executing precise algorithmic graph traversals without hallucinating edges or conflating distinct semantic nodes \citep{jiang2025reasoning, kiguchi2025multi}.

To formalize this representational gap, we define the geometric boundary of Topological Collapse as a direct, bounded consequence of the Impedance Mismatch ($\mathcal{I}$) established in Section~\ref{sec:formalization}. When mapping the discrete metric space of the graph $\mathcal{K} = (\mathcal{V}, \mathcal{E})$ into the continuous latent space $\mathcal{M}$ via an embedding function $f$, the structural distortion cannot be arbitrarily minimized. 

According to Bourgain's Embedding Theorem, embedding a finite metric space of $|\mathcal{V}|$ points into a Euclidean space inherently introduces a minimum structural distortion mathematically bounded by $\Omega(\log |\mathcal{V}|)$. Therefore, we can formally bound the Impedance Mismatch for Level 2 integrations as $\mathcal{I} \ge \Omega(\log |\mathcal{V}|)$. As the size of the ontology grows, this minimum distortion grows logarithmically. Because a perfect, distance-preserving semantic alignment strictly requires $\mathcal{I} = 1$, achieving zero-distortion integration at the embedding level is mathematically impossible. The continuous vector space natively lacks the geometric capacity to preserve the discrete graph structure, unavoidably forcing distinct semantic nodes to overlap and destroying the boundaries required for precise algorithmic traversals.

\subsection{Level 3: Architectural Integration (Attention-Level)}

The most advanced frontier of current research involves directly modifying the internal computational routing of the transformer architecture to explicitly accommodate graph structures. Rather than treating the knowledge graph as an external text payload or an aligned input vector, these methodologies inject graph priors directly into the message-passing framework or the self-attention calculations of the model \citep{luo2025integrating, yasunaga2021qa}. Recent architectural innovations include Graph-Guided Attention modules that non-invasively rewire the native attention matrices of the foundational model based strictly on knowledge graph adjacency \citep{zhang2025enhancing, zhai2026knowledgefusionbidirectionalinformation}. Parallel frameworks utilize cross-attention mechanisms to inject semantic graph prompts dynamically across intermediate hidden layers \citep{hu2022knowledgeable}.

Critique: While architecturally integrated models exhibit state-of-the-art empirical performance on complex reasoning benchmarks \citep{wang2024large, yasunaga2021qa}, they remain theoretically incomplete. They are computationally expensive to scale. They still treat the knowledge graph as an externalized constraint that must be dynamically consulted rather than functioning as an internalized, native knowledge structure. The fundamental mathematical friction remains unresolved because the neural network is still relying on continuous attention weights to approximate discrete logical routing \citep{pan2024unifying, luo2025integrating}. Until the underlying transformer architecture natively supports discontinuous structural subspaces within its residual stream, true semantic fusion will remain out of reach \citep{zhai2026knowledgefusionbidirectionalinformation}.

To mathematically formalize this architectural limitation, we define the boundary of Attention Approximation Leakage. In a pure symbolic system, logical routing is executed via a discrete adjacency matrix $A \in \{0, 1\}^{n \times n}$. Architecturally integrated foundational models attempt to approximate this discrete routing using continuous attention matrices $A_{\text{soft}} \in (0, 1)^{n \times n}$. 

Because the standard attention mechanism relies on the softmax function, it strictly outputs positive probabilities. Approximating a hard, discrete zero (indicating no relationship) requires infinite negative logits, which is impossible in a stable training regime. Therefore, every non-adjacent node contributes a strictly positive residual leakage error $\delta > 0$ during the message-passing calculation. When the model attempts to execute a multi-hop logical query of depth $k$, the routing calculation approximates $(A_{\text{soft}})^k$. As $k$ increases, the continuous leakage error $\delta$ compounds exponentially, leading to severe representation over-smoothing. The precise signal of the true discrete reasoning path is inevitably drowned out by the accumulated noise of the continuous space, proving that approximating discrete routing with continuous attention weights is mathematically unsustainable for deep logical deduction.

\begin{table*}[t]
\centering
\small
\renewcommand{\arraystretch}{1.4}
\begin{tabular}{|l|p{2cm}|p{4.5cm}|p{4.5cm}|}
\hline
\textbf{Integration Level} & \textbf{Mechanism} & \textbf{Formal Mathematical Bottleneck} & \textbf{Asymptotic Failure Mode} \\ \hline

Level 1: Surface & Lexical Prompt Injection & Lexical Bottleneck: \newline $\mathcal{O}(b^k) > L$ & Context truncation; inability to encode exponential path complexity. \\ \hline

Level 2: Embedding & Latent Vector Alignment & Topological Collapse: \newline $D(f) \ge \Omega(\log |\mathcal{V}|)$ & Semantic conflation; distortion of discrete relational boundaries. \\ \hline

Level 3: Architecture & Graph-Guided Attention & Approximation Leakage: \newline Compounding softmax error $\delta$ in $(A_{\text{soft}})^k$ & Representation over-smoothing; discrete signal drowned in continuous noise. \\ \hline
\end{tabular}

\caption{A theoretical taxonomy of neuro-symbolic integration strategies, classified by their fundamental mathematical bottlenecks and asymptotic failure modes during multi-hop reasoning.}
\label{tab:integration_limits}
\end{table*}

\section{Core Bottlenecks Preventing True Fusion}
\label{sec:bottlenecks}

To move past the design limits of current integration strategies and achieve true semantic fusion, the community must address three fundamental bottlenecks. These barriers represent incompatibilities between discrete structural constraints and continuous latent spaces.

\subsection{Bottleneck A: The Curse of Differentiable Logic}

A prevalent method for injecting discrete logic into continuous models utilizes differentiable logic frameworks, which relax Boolean connectives and quantifiers into continuous operators \citep{rocktaschel2017end, evans2018learning, van2022analyzing}. Soft relaxations algorithmically map strict truth values to the continuous interval $[0, 1]$ via t-norms, s-norms, and fuzzy aggregation operators \citep{van2022analyzing, manhaeve2018deep}. However, this mapping introduces an optimization bottleneck. The resulting loss landscapes are non-linear and suffer from acute gradient saturation \citep{giunchiglia2022deep, wang2019satnet}. Once a logical formula is nearly satisfied, the gradients vanish entirely, prematurely halting the optimization process before true semantic alignment is achieved \citep{van2022analyzing, minervini2020differentiable}. 

Furthermore, soft truth values break classical logical equivalences. In a discrete knowledge graph, De Morgan's laws and contraposition hold absolute certainty. In a relaxed tensor space, these functionally equivalent symbolic rules often yield entirely divergent optimization paths \citep{giunchiglia2022deep, wang2019satnet}. This inherent conflict makes robust constraint satisfaction mathematically unstable under stochastic gradient descent. Consequently, researchers are forced to choose between Boolean faithfulness and optimization amenability \citep{van2022analyzing, garcez2022neural}.

\subsection{Bottleneck B: Structural and Geometric Interference}

The second barrier is structural and geometric interference. In a discrete graph, edges provide perfect relational insulation. Editing the relation between a subject node and an object node has no impact on adjacent graph edges. In a continuous representation space, such absolute geometric isolation is mathematically impossible \citep{meng2022locating, elhage2021mathematical}. When discrete symbolic structures are encoded into high-dimensional vectors, they overlap and blend within the same dense space \citep{elhage2021mathematical}. 

Updating parametric memory to modify a specific bound relation inherently warps the local geometry of the embedding representation space \citep{meng2022locating, hase2024does}. As the number of overlapping facts in the residual stream increases, theoretical capacity limits are reached, and knowledge extraction operations inevitably suffer from catastrophic crosstalk \citep{yao2023editing, zhong2023mquake}. Surgically editing a specific semantic relation can inadvertently alter adjacent, structurally unrelated knowledge \citep{meng2022locating, de2021editing, cohen2023evaluating}. The fluid nature of the transformer's residual stream lacks the strict orthogonality required to perfectly insulate discrete variables during continuous updates \citep{wang2024knowledge}. This leads to the logical consistency breaking down entirely under minor parameter perturbations \citep{cohen2023evaluating, zhong2023mquake, hase2024does}.

\subsection{Bottleneck C: The Symbol Grounding Asymmetry}

The final bottleneck centers on the asymmetry in symbol grounding \citep{harnad1990symbol, ji2021survey}. Knowledge graphs rely on unique entity identifiers to maintain strict referential integrity across diverse contexts \citep{hogan2021knowledge}. On the other hand, foundational models process information through contextualized, distributed sub-word token representations \citep{brown2020language, openai2024gpt4}. 

Aligning abstract, immutable symbols with fluid data patterns remains a major theoretical challenge \citep{pan2024unifying, sun2024large}. While prior works attempt to bridge this gap using contrastive alignment or dedicated entity embeddings, these methods assume a static mapping that ignores the dynamically overlapping nature of language models \citep{pan2024unifying, luo2025integrating, zhang2025enhancing}. Natively integrating symbolic knowledge requires a mechanism to dynamically instantiate and bind discrete roles to continuous fillers without losing the strict identity of the original symbol \citep{garcez2022neural, smolensky1990tensor}. Until this structural asymmetry is mathematically resolved, hybrid models will continue to rely on shallow pattern matching rather than exhibiting true, provable compositional generalization \citep{lake2017building, bahdanau2019systematic, ruis2020benchmark}.

\section{A Roadmap for the Knowledge Lifecycle}
\label{sec:roadmap}
To resolve the bottlenecks in Section \ref{sec:bottlenecks} and the impedance mismatch, we build upon the framework established by \citep{dhayalkar_roadmap} to propose an actionable three-stage knowledge lifecycle roadmap that transcends lexical bridging.

\subsection{Emergence (Pre-training): Structured Residual Streams}

Current pre-training paradigms rely on unconstrained geometric optimization. This reliance directly causes the structural and geometric interference of factual knowledge observed during complex reasoning tasks \citep{elhage2021mathematical, bricken2023towards}. However, recent breakthroughs in Representation Engineering demonstrate that high-level concepts naturally manifest as stable subspace directions or principal-eigenvector backbones within the transformer's residual stream \citep{zou2023representation, park2023linear}. Furthermore, models can natively recover spatial separations that directly map to structured human concept categories \citep{wang2024concept, li2023emergent}. 

To formalize this phenomenon, we propose the architectural development of \textit{Structured Residual Streams}. Rather than allowing facts to overlap arbitrarily across the entire embedding latent space, future architectures should introduce explicit graph-theoretic inductive biases during pre-training \citep{pan2024unifying, luo2025integrating}. By applying regularization penalties that enforce orthogonal subspaces for distinct knowledge domains, discrete relational structures could emerge natively within the continuous weights. This would equip the model with an inherent, mathematically insulated structure, preventing the catastrophic crosstalk that currently degrades multi-hop reasoning \citep{frady2023variable}.

\subsection{Injection (Inference): Latent Sub-graph Injection via VSAs}

The industry standard of text-based retrieval is limited by tokenization bottlenecks and the high influence of the continuous parametric prior \citep{mallen2023trust, lewis2020retrieval}. To bypass this, we must shift from external lexical prompting to \textit{Latent Sub-graph Injection}. We propose utilizing Vector Symbolic Architectures (VSAs) as the mathematical bridge to achieve this integration natively.

VSAs provide a well-defined algebraic framework using operations like binding, bundling, and permutation to represent complex discrete graph data within unified high-dimensional vector spaces \citep{kanerva2009hyperdimensional, kleyko2022vector}. VSAs retain fixed-dimensional vectors that align naturally with the native embeddings of the standard transformer architecture \citep{smolensky1990tensor}. By encoding a retrieved knowledge graph subgraph directly into a VSA hypervector, researchers can inject explicit role-filler bindings directly into the intermediate attention layers of the foundation model at inference time \citep{meng2022locating, kanerva2009hyperdimensional, dhayalkar_attention}. This bypasses the superficial text layer and forces the model to condition its generation on strict, mathematically bound relations rather than probabilistic text prompts.

\subsection{Updating (Editing): Orthogonal Subspace Editing}

The editability conflict requires a new mathematical approach to model updates. Current continuous knowledge editing regimes suffer from a performance decline in knowledge retention as sequential edits increase \citep{meng2022locating, mitchell2022fast, de2021editing}. While recent methods have advanced the ability to update long-form knowledge using dynamic weight adjustments, they still grapple with coupling of the continuous vector space \citep{yao2023editing, zhong2023mquake}. 

To guarantee localized factual updates without neighborhood interference, we call for the formalization of \textit{Orthogonal Subspace Editing}. Recent dissections of perturbation weights indicate that disentangled and sparsified knowledge representations can alleviate performance degradation during continuous editing \citep{hase2024does}. Building on this insight, we hypothesize that by 
projecting targeted factual edits strictly along orthogonal feature directions that do not activate unrelated semantic concepts, we can achieve updates that are mathematically equivalent to localized edge-insertion. This theoretical direction would allow foundational models to be patched dynamically and safely, finally bringing the reliable editability of symbolic knowledge bases to neural parameter spaces \citep{pan2024unifying, luo2025integrating, meng2022locating}.

\section{Conclusion}
\label{sec:conclusion}

Continuing to treat knowledge graphs merely as external databases or retrieval dictionaries fundamentally limits the evolutionary trajectory of foundation models. Throughout this paper, we have demonstrated that the current industrial standard of text-based retrieval acts only as a superficial patch over a much deeper structural divide. We defined this divide as the Impedance Mismatch, a mathematical friction that occurs when attempting to force rigid, deterministic graph relational structures into fluid, probabilistic embedding spaces. 

By categorizing existing integration attempts into a hierarchy of maturity, we revealed that neither lexical prompt injection nor continuous representation alignment can preserve the strict logical motifs required for reliable, multi-hop reasoning. The true barriers to semantic fusion are not engineering hurdles, but rather deep theoretical bottlenecks. The saturation of differentiable logic, the structural and geometric interference of continuous memory, and the fundamental asymmetry of symbol grounding collectively prevent standard transformer architectures from natively internalizing discrete symbolic structures.

To construct truly knowledgeable foundation models, the research community must move beyond the paradigm of lexical bridging. We must confront the fundamental mathematical friction between discrete certainty and continuous probability directly at the architectural level. By pursuing structured residual streams, latent sub-graph injection via vector-symbolic architectures, and orthogonal subspace editing, we can transition from models that mimic factual recall to systems that genuinely harbor structured, editable knowledge. Resolving this impedance mismatch is the necessary next step in the knowledge lifecycle, enabling a future where the precision of symbolic logic and the expressivity of parametric memory are seamlessly and mathematically fused.

\section*{Limitations}
While this paper establishes a rigorous mathematical foundation for neuro-symbolic integration, it focuses strictly on formal analysis and does not include empirical experiments. Consequently, our proposed frameworks currently serve as theoretical blueprints. Bridging these formalisms, such as Structured Residual Streams and VSA injection into scalable training regimes, represents a natural next step for empirical research. Additionally, because our models assume perfectly deterministic knowledge graphs, future work must explore how these strict geometric constraints adapt to the noise and contradictions inherent in real-world knowledge bases.

\bibliography{custom}

@inproceedings{vaswani,
 author = {Vaswani, Ashish and Shazeer, Noam and Parmar, Niki and Uszkoreit, Jakob and Jones, Llion and Gomez, Aidan N and Kaiser, \L ukasz and Polosukhin, Illia},
 booktitle = {Advances in Neural Information Processing Systems},
 editor = {I. Guyon and U. Von Luxburg and S. Bengio and H. Wallach and R. Fergus and S. Vishwanathan and R. Garnett},
 pages = {},
 publisher = {Curran Associates, Inc.},
 title = {Attention is All you Need},
 url = {https://proceedings.neurips.cc/paper_files/paper/2017/file/3f5ee243547dee91fbd053c1c4a845aa-Paper.pdf},
 volume = {30},
 year = {2017}
}

@article{pan2024unifying,
   title={Unifying Large Language Models and Knowledge Graphs: A Roadmap},
   volume={36},
   ISSN={2326-3865},
   url={http://dx.doi.org/10.1109/TKDE.2024.3352100},
   DOI={10.1109/tkde.2024.3352100},
   number={7},
   journal={IEEE Transactions on Knowledge and Data Engineering},
   publisher={Institute of Electrical and Electronics Engineers (IEEE)},
   author={Pan, Shirui and Luo, Linhao and Wang, Yufei and Chen, Chen and Wang, Jiapu and Wu, Xindong},
   year={2024},
   month=jul, pages={3580–3599} }

@article{luo2025integrating,
  title={Integrating Large Language Models and Knowledge Graphs for Next-level AGI},
  author={Linhao Luo and Carl Yang and Evgeny Kharlamov and Shirui Pan},
  journal={Companion Proceedings of the ACM on Web Conference 2025},
  year={2025},
  url={https://api.semanticscholar.org/CorpusID:277057192}
}

@article{ji2021survey,
   title={A Survey on Knowledge Graphs: Representation, Acquisition, and Applications},
   volume={33},
   ISSN={2162-2388},
   url={http://dx.doi.org/10.1109/TNNLS.2021.3070843},
   DOI={10.1109/tnnls.2021.3070843},
   number={2},
   journal={IEEE Transactions on Neural Networks and Learning Systems},
   publisher={Institute of Electrical and Electronics Engineers (IEEE)},
   author={Ji, Shaoxiong and Pan, Shirui and Cambria, Erik and Marttinen, Pekka and Yu, Philip S.},
   year={2022},
   month=feb, pages={494–514} }

@article{hogan2021knowledge,
   title={Knowledge Graphs},
   volume={54},
   ISSN={1557-7341},
   url={http://dx.doi.org/10.1145/3447772},
   DOI={10.1145/3447772},
   number={4},
   journal={ACM Computing Surveys},
   publisher={Association for Computing Machinery (ACM)},
   author={Hogan, Aidan and Blomqvist, Eva and Cochez, Michael and D’amato, Claudia and Melo, Gerard De and Gutierrez, Claudio and Kirrane, Sabrina and Gayo, José Emilio Labra and Navigli, Roberto and Neumaier, Sebastian and Ngomo, Axel-Cyrille Ngonga and Polleres, Axel and Rashid, Sabbir M. and Rula, Anisa and Schmelzeisen, Lukas and Sequeda, Juan and Staab, Steffen and Zimmermann, Antoine},
   year={2021},
   month=jul, pages={1–37} }

@inproceedings{brown2020language,
 author = {Brown, Tom and Mann, Benjamin and Ryder, Nick and Subbiah, Melanie and Kaplan, Jared D and Dhariwal, Prafulla and Neelakantan, Arvind and Shyam, Pranav and Sastry, Girish and Askell, Amanda and Agarwal, Sandhini and Herbert-Voss, Ariel and Krueger, Gretchen and Henighan, Tom and Child, Rewon and Ramesh, Aditya and Ziegler, Daniel and Wu, Jeffrey and Winter, Clemens and Hesse, Chris and Chen, Mark and Sigler, Eric and Litwin, Mateusz and Gray, Scott and Chess, Benjamin and Clark, Jack and Berner, Christopher and McCandlish, Sam and Radford, Alec and Sutskever, Ilya and Amodei, Dario},
 booktitle = {Advances in Neural Information Processing Systems},
 editor = {H. Larochelle and M. Ranzato and R. Hadsell and M.F. Balcan and H. Lin},
 pages = {1877--1901},
 publisher = {Curran Associates, Inc.},
 title = {Language Models are Few-Shot Learners},
 url = {https://proceedings.neurips.cc/paper_files/paper/2020/file/1457c0d6bfcb4967418bfb8ac142f64a-Paper.pdf},
 volume = {33},
 year = {2020}
}

@misc{touvron2023llama,
      title={LLaMA: Open and Efficient Foundation Language Models}, 
      author={Hugo Touvron and Thibaut Lavril and Gautier Izacard and Xavier Martinet and Marie-Anne Lachaux and Timothée Lacroix and Baptiste Rozière and Naman Goyal and Eric Hambro and Faisal Azhar and Aurelien Rodriguez and Armand Joulin and Edouard Grave and Guillaume Lample},
      year={2023},
      eprint={2302.13971},
      archivePrefix={arXiv},
      primaryClass={cs.CL},
      url={https://arxiv.org/abs/2302.13971}, 
}

@misc{openai2024gpt4,
      title={GPT-4 Technical Report}, 
      author={OpenAI and Josh Achiam and Steven Adler and Sandhini Agarwal and Lama Ahmad and Ilge Akkaya and Florencia Leoni Aleman and Diogo Almeida and Janko Altenschmidt and Sam Altman and Shyamal Anadkat and Red Avila and Igor Babuschkin and Suchir Balaji and Valerie Balcom and Paul Baltescu and Haiming Bao and Mohammad Bavarian and Jeff Belgum and Irwan Bello and Jake Berdine and Gabriel Bernadett-Shapiro and Christopher Berner and Lenny Bogdonoff and Oleg Boiko and Madelaine Boyd and Anna-Luisa Brakman and Greg Brockman and Tim Brooks and Miles Brundage and Kevin Button and Trevor Cai and Rosie Campbell and Andrew Cann and Brittany Carey and Chelsea Carlson and Rory Carmichael and Brooke Chan and Che Chang and Fotis Chantzis and Derek Chen and Sully Chen and Ruby Chen and Jason Chen and Mark Chen and Ben Chess and Chester Cho and Casey Chu and Hyung Won Chung and Dave Cummings and Jeremiah Currier and Yunxing Dai and Cory Decareaux and Thomas Degry and Noah Deutsch and Damien Deville and Arka Dhar and David Dohan and Steve Dowling and Sheila Dunning and Adrien Ecoffet and Atty Eleti and Tyna Eloundou and David Farhi and Liam Fedus and Niko Felix and Simón Posada Fishman and Juston Forte and Isabella Fulford and Leo Gao and Elie Georges and Christian Gibson and Vik Goel and Tarun Gogineni and Gabriel Goh and Rapha Gontijo-Lopes and Jonathan Gordon and Morgan Grafstein and Scott Gray and Ryan Greene and Joshua Gross and Shixiang Shane Gu and Yufei Guo and Chris Hallacy and Jesse Han and Jeff Harris and Yuchen He and Mike Heaton and Johannes Heidecke and Chris Hesse and Alan Hickey and Wade Hickey and Peter Hoeschele and Brandon Houghton and Kenny Hsu and Shengli Hu and Xin Hu and Joost Huizinga and Shantanu Jain and Shawn Jain and Joanne Jang and Angela Jiang and Roger Jiang and Haozhun Jin and Denny Jin and Shino Jomoto and Billie Jonn and Heewoo Jun and Tomer Kaftan and Łukasz Kaiser and Ali Kamali and Ingmar Kanitscheider and Nitish Shirish Keskar and Tabarak Khan and Logan Kilpatrick and Jong Wook Kim and Christina Kim and Yongjik Kim and Jan Hendrik Kirchner and Jamie Kiros and Matt Knight and Daniel Kokotajlo and Łukasz Kondraciuk and Andrew Kondrich and Aris Konstantinidis and Kyle Kosic and Gretchen Krueger and Vishal Kuo and Michael Lampe and Ikai Lan and Teddy Lee and Jan Leike and Jade Leung and Daniel Levy and Chak Ming Li and Rachel Lim and Molly Lin and Stephanie Lin and Mateusz Litwin and Theresa Lopez and Ryan Lowe and Patricia Lue and Anna Makanju and Kim Malfacini and Sam Manning and Todor Markov and Yaniv Markovski and Bianca Martin and Katie Mayer and Andrew Mayne and Bob McGrew and Scott Mayer McKinney and Christine McLeavey and Paul McMillan and Jake McNeil and David Medina and Aalok Mehta and Jacob Menick and Luke Metz and Andrey Mishchenko and Pamela Mishkin and Vinnie Monaco and Evan Morikawa and Daniel Mossing and Tong Mu and Mira Murati and Oleg Murk and David Mély and Ashvin Nair and Reiichiro Nakano and Rajeev Nayak and Arvind Neelakantan and Richard Ngo and Hyeonwoo Noh and Long Ouyang and Cullen O'Keefe and Jakub Pachocki and Alex Paino and Joe Palermo and Ashley Pantuliano and Giambattista Parascandolo and Joel Parish and Emy Parparita and Alex Passos and Mikhail Pavlov and Andrew Peng and Adam Perelman and Filipe de Avila Belbute Peres and Michael Petrov and Henrique Ponde de Oliveira Pinto and Michael and Pokorny and Michelle Pokrass and Vitchyr H. Pong and Tolly Powell and Alethea Power and Boris Power and Elizabeth Proehl and Raul Puri and Alec Radford and Jack Rae and Aditya Ramesh and Cameron Raymond and Francis Real and Kendra Rimbach and Carl Ross and Bob Rotsted and Henri Roussez and Nick Ryder and Mario Saltarelli and Ted Sanders and Shibani Santurkar and Girish Sastry and Heather Schmidt and David Schnurr and John Schulman and Daniel Selsam and Kyla Sheppard and Toki Sherbakov and Jessica Shieh and Sarah Shoker and Pranav Shyam and Szymon Sidor and Eric Sigler and Maddie Simens and Jordan Sitkin and Katarina Slama and Ian Sohl and Benjamin Sokolowsky and Yang Song and Natalie Staudacher and Felipe Petroski Such and Natalie Summers and Ilya Sutskever and Jie Tang and Nikolas Tezak and Madeleine B. Thompson and Phil Tillet and Amin Tootoonchian and Elizabeth Tseng and Preston Tuggle and Nick Turley and Jerry Tworek and Juan Felipe Cerón Uribe and Andrea Vallone and Arun Vijayvergiya and Chelsea Voss and Carroll Wainwright and Justin Jay Wang and Alvin Wang and Ben Wang and Jonathan Ward and Jason Wei and CJ Weinmann and Akila Welihinda and Peter Welinder and Jiayi Weng and Lilian Weng and Matt Wiethoff and Dave Willner and Clemens Winter and Samuel Wolrich and Hannah Wong and Lauren Workman and Sherwin Wu and Jeff Wu and Michael Wu and Kai Xiao and Tao Xu and Sarah Yoo and Kevin Yu and Qiming Yuan and Wojciech Zaremba and Rowan Zellers and Chong Zhang and Marvin Zhang and Shengjia Zhao and Tianhao Zheng and Juntang Zhuang and William Zhuk and Barret Zoph},
      year={2024},
      eprint={2303.08774},
      archivePrefix={arXiv},
      primaryClass={cs.CL},
      url={https://arxiv.org/abs/2303.08774}, 
}

@inproceedings{li2026survey,
  author    = {Xu, Ran and Jiang, Patrick and Luo, Linhao and Xiao, Cao and Cross, Adam and Pan, Shirui and Sun, Jimeng and Yang, Carl},
  title     = {A Survey on Unifying Large Language Models and Knowledge Graphs for Biomedicine and Healthcare},
  booktitle = {Proceedings of the 31st ACM SIGKDD Conference on Knowledge Discovery and Data Mining (KDD '25)},
  year      = {2025},
  month     = {August},
  pages     = {6195--6205},
  doi       = {10.1145/3711896.3736556},
  note      = {PMID: 41858611; PMCID: PMC12995553}
}

@inproceedings{ma2025large,
    title = "Large Language Models Meet Knowledge Graphs for Question Answering: Synthesis and Opportunities",
    author = "Ma, Chuangtao  and
      Chen, Yongrui  and
      Wu, Tianxing  and
      Khan, Arijit  and
      Wang, Haofen",
    editor = "Christodoulopoulos, Christos  and
      Chakraborty, Tanmoy  and
      Rose, Carolyn  and
      Peng, Violet",
    booktitle = "Proceedings of the 2025 Conference on Empirical Methods in Natural Language Processing",
    month = nov,
    year = "2025",
    address = "Suzhou, China",
    publisher = "Association for Computational Linguistics",
    url = "https://aclanthology.org/2025.emnlp-main.1249/",
    doi = "10.18653/v1/2025.emnlp-main.1249",
    pages = "24578--24597",
    ISBN = "979-8-89176-332-6"
}

@inproceedings{lewis2020retrieval,
 author = {Lewis, Patrick and Perez, Ethan and Piktus, Aleksandra and Petroni, Fabio and Karpukhin, Vladimir and Goyal, Naman and K\"{u}ttler, Heinrich and Lewis, Mike and Yih, Wen-tau and Rockt\"{a}schel, Tim and Riedel, Sebastian and Kiela, Douwe},
 booktitle = {Advances in Neural Information Processing Systems},
 editor = {H. Larochelle and M. Ranzato and R. Hadsell and M.F. Balcan and H. Lin},
 pages = {9459--9474},
 publisher = {Curran Associates, Inc.},
 title = {Retrieval-Augmented Generation for Knowledge-Intensive NLP Tasks},
 url = {https://proceedings.neurips.cc/paper_files/paper/2020/file/6b493230205f780e1bc26945df7481e5-Paper.pdf},
 volume = {33},
 year = {2020}
}

@misc{guu2020realm,
      title={REALM: Retrieval-Augmented Language Model Pre-Training}, 
      author={Kelvin Guu and Kenton Lee and Zora Tung and Panupong Pasupat and Ming-Wei Chang},
      year={2020},
      eprint={2002.08909},
      archivePrefix={arXiv},
      primaryClass={cs.CL},
      url={https://arxiv.org/abs/2002.08909}, 
}

@misc{gao2023retrieval,
      title={Retrieval-Augmented Generation for Large Language Models: A Survey}, 
      author={Yunfan Gao and Yun Xiong and Xinyu Gao and Kangxiang Jia and Jinliu Pan and Yuxi Bi and Yi Dai and Jiawei Sun and Meng Wang and Haofen Wang},
      year={2024},
      eprint={2312.10997},
      archivePrefix={arXiv},
      primaryClass={cs.CL},
      url={https://arxiv.org/abs/2312.10997}, 
}

@misc{edge2024graphrag,
      title={From Local to Global: A Graph RAG Approach to Query-Focused Summarization}, 
      author={Darren Edge and Ha Trinh and Newman Cheng and Joshua Bradley and Alex Chao and Apurva Mody and Steven Truitt and Dasha Metropolitansky and Robert Osazuwa Ness and Jonathan Larson},
      year={2025},
      eprint={2404.16130},
      archivePrefix={arXiv},
      primaryClass={cs.CL},
      url={https://arxiv.org/abs/2404.16130}, 
}

@inproceedings{xu2024retrieval, series={SIGIR 2024},
   title={Retrieval-Augmented Generation with Knowledge Graphs for Customer Service Question Answering},
   url={http://dx.doi.org/10.1145/3626772.3661370},
   DOI={10.1145/3626772.3661370},
   booktitle={Proceedings of the 47th International ACM SIGIR Conference on Research and Development in Information Retrieval},
   publisher={ACM},
   author={Xu, Zhentao and Cruz, Mark Jerome and Guevara, Matthew and Wang, Tie and Deshpande, Manasi and Wang, Xiaofeng and Li, Zheng},
   year={2024},
   month=jul, pages={2905–2909},
   collection={SIGIR 2024} }

@misc{chen2025practices,
      title={LLM-empowered knowledge graph construction: A survey}, 
      author={Haonan Bian},
      year={2025},
      eprint={2510.20345},
      archivePrefix={arXiv},
      primaryClass={cs.AI},
      url={https://arxiv.org/abs/2510.20345}, 
}

@misc{wang2024large,
      title={Large Language Models on Graphs: A Comprehensive Survey}, 
      author={Bowen Jin and Gang Liu and Chi Han and Meng Jiang and Heng Ji and Jiawei Han},
      year={2024},
      eprint={2312.02783},
      archivePrefix={arXiv},
      primaryClass={cs.CL},
      url={https://arxiv.org/abs/2312.02783}, 
}

@inproceedings{regraphrag2025,
    title = "{R}e{G}raph{RAG}: Reorganizing Fragmented Knowledge Graphs for Multi-Perspective Retrieval-Augmented Generation",
    author = "Kim, Soohyeong  and
      Hwang, Seok Jun  and
      Kim, JungHyoun  and
      Park, Jeonghyeon  and
      Choi, Yong Suk",
    editor = "Christodoulopoulos, Christos  and
      Chakraborty, Tanmoy  and
      Rose, Carolyn  and
      Peng, Violet",
    booktitle = "Findings of the Association for Computational Linguistics: EMNLP 2025",
    month = nov,
    year = "2025",
    address = "Suzhou, China",
    publisher = "Association for Computational Linguistics",
    url = "https://aclanthology.org/2025.findings-emnlp.290/",
    doi = "10.18653/v1/2025.findings-emnlp.290",
    pages = "5426--5443",
    ISBN = "979-8-89176-335-7"
}

@InProceedings{jiang2025reasoning,
  title = 	 {Graph-constrained Reasoning: Faithful Reasoning on Knowledge Graphs with Large Language Models},
  author =       {Luo, Linhao and Zhao, Zicheng and Haffari, Gholamreza and Li, Yuan-Fang and Gong, Chen and Pan, Shirui},
  booktitle = 	 {Proceedings of the 42nd International Conference on Machine Learning},
  pages = 	 {41540--41565},
  year = 	 {2025},
  editor = 	 {Singh, Aarti and Fazel, Maryam and Hsu, Daniel and Lacoste-Julien, Simon and Berkenkamp, Felix and Maharaj, Tegan and Wagstaff, Kiri and Zhu, Jerry},
  volume = 	 {267},
  series = 	 {Proceedings of Machine Learning Research},
  month = 	 {13--19 Jul},
  publisher =    {PMLR},
  url = 	 {https://proceedings.mlr.press/v267/luo25t.html}
}

@article{harnad1990symbol,
author = {Harnad, Stevan},
year = {1990},
month = {01},
pages = {335-346},
title = {Harnad, S. (1990). The symbol grounding problem. Physica D: Nonlinear Phenomena, 42(1-3), 335-346.},
volume = {42}
}

@article{vankrieken2022analyzing,
   title={Analyzing Differentiable Fuzzy Logic Operators},
   volume={302},
   ISSN={0004-3702},
   url={http://dx.doi.org/10.1016/j.artint.2021.103602},
   DOI={10.1016/j.artint.2021.103602},
   journal={Artificial Intelligence},
   publisher={Elsevier BV},
   author={van Krieken, Emile and Acar, Erman and van Harmelen, Frank},
   year={2022},
   month=jan, pages={103602} }

@inproceedings{
neurips2025synergizing,
title={Synergizing Large Language Models and Knowledge Graphs in Science: A Survey},
author={Zhihui Zhu and Yuqi Tang and Qiang Zhang and Keyan Ding},
booktitle={NeurIPS 2025 AI for Science Workshop},
year={2025},
url={https://openreview.net/forum?id=WUFfhhHNsz}
}

@misc{garcez2022neural,
      title={Neural-Symbolic Computing: An Effective Methodology for Principled Integration of Machine Learning and Reasoning}, 
      author={Artur d'Avila Garcez and Marco Gori and Luis C. Lamb and Luciano Serafini and Michael Spranger and Son N. Tran},
      year={2019},
      eprint={1905.06088},
      archivePrefix={arXiv},
      primaryClass={cs.AI},
      url={https://arxiv.org/abs/1905.06088}, 
}

@misc{zhang2025enhancing,
      title={Enhancing Large Language Models with Reliable Knowledge Graphs}, 
      author={Qinggang Zhang},
      year={2025},
      eprint={2506.13178},
      archivePrefix={arXiv},
      primaryClass={cs.CL},
      url={https://arxiv.org/abs/2506.13178}, 
}

@inproceedings{yao2023editing,
    title = "Editing Large Language Models: Problems, Methods, and Opportunities",
    author = "Yao, Yunzhi  and
      Wang, Peng  and
      Tian, Bozhong  and
      Cheng, Siyuan  and
      Li, Zhoubo  and
      Deng, Shumin  and
      Chen, Huajun  and
      Zhang, Ningyu",
    editor = "Bouamor, Houda  and
      Pino, Juan  and
      Bali, Kalika",
    booktitle = "Proceedings of the 2023 Conference on Empirical Methods in Natural Language Processing",
    month = dec,
    year = "2023",
    address = "Singapore",
    publisher = "Association for Computational Linguistics",
    url = "https://aclanthology.org/2023.emnlp-main.632/",
    doi = "10.18653/v1/2023.emnlp-main.632",
    pages = "10222--10240"
}

@inproceedings{meng2022locating,
author = {Meng, Kevin and Bau, David and Andonian, Alex and Belinkov, Yonatan},
title = {Locating and editing factual associations in GPT},
year = {2022},
isbn = {9781713871088},
publisher = {Curran Associates Inc.},
address = {Red Hook, NY, USA},
booktitle = {Proceedings of the 36th International Conference on Neural Information Processing Systems},
articleno = {1262},
numpages = {14},
location = {New Orleans, LA, USA},
series = {NIPS '22}
}

@misc{mitchell2022fast,
      title={Fast Model Editing at Scale}, 
      author={Eric Mitchell and Charles Lin and Antoine Bosselut and Chelsea Finn and Christopher D. Manning},
      year={2022},
      eprint={2110.11309},
      archivePrefix={arXiv},
      primaryClass={cs.LG},
      url={https://arxiv.org/abs/2110.11309}, 
}

@inproceedings{de2021editing,
    title = "Editing Factual Knowledge in Language Models",
    author = "De Cao, Nicola  and
      Aziz, Wilker  and
      Titov, Ivan",
    editor = "Moens, Marie-Francine  and
      Huang, Xuanjing  and
      Specia, Lucia  and
      Yih, Scott Wen-tau",
    booktitle = "Proceedings of the 2021 Conference on Empirical Methods in Natural Language Processing",
    month = nov,
    year = "2021",
    address = "Online and Punta Cana, Dominican Republic",
    publisher = "Association for Computational Linguistics",
    url = "https://aclanthology.org/2021.emnlp-main.522/",
    doi = "10.18653/v1/2021.emnlp-main.522",
    pages = "6491--6506"
}

@article{
dhayalkar_roadmap,
author = {Sahil Rajesh Dhayalkar },
title = {Neuro-Symbolic Reasoning: A Roadmap of Unsolved Core Questions},
journal = {TechRxiv},
volume = {2025},
number = {1210},
pages = {},
year = {2025},
doi = {10.36227/techrxiv.176539555.52683902/v1},
URL = {https://www.techrxiv.org/doi/abs/10.36227/techrxiv.176539555.52683902/v1},
eprint = {https://www.techrxiv.org/doi/pdf/10.36227/techrxiv.176539555.52683902/v1}}

@inproceedings{
hase2024does,
title={Does Localization Inform Editing? Surprising Differences in Causality-Based Localization vs. Knowledge Editing in Language Models},
author={Peter Hase and Mohit Bansal and Been Kim and Asma Ghandeharioun},
booktitle={Thirty-seventh Conference on Neural Information Processing Systems},
year={2023},
url={https://openreview.net/forum?id=EldbUlZtbd}
}

@misc{chen2025gril,
      title={GRIL: Knowledge Graph Retrieval-Integrated Learning with Large Language Models}, 
      author={Jialin Chen and Houyu Zhang and Seongjun Yun and Alejandro Mottini and Rex Ying and Xiang Song and Vassilis N. Ioannidis and Zheng Li and Qingjun Cui},
      year={2025},
      eprint={2509.16502},
      archivePrefix={arXiv},
      primaryClass={cs.LG},
      url={https://arxiv.org/abs/2509.16502}, 
}

@inproceedings{liu2025hyp,
    author       = {Liu, Zhaotai and Sack, Harald and Gesese, Genet Asefa},
    year         = {2025},
    title        = {HyP-KGRAG: Hypothetical Path-Based Knowledge Graph Retrieval Augmented Generation with DeepSeek},
    pages        = {45 - 55},
    eventtitle   = {2nd RAGE-KG: The International Workshop on Retrieval-Augmented Generation Enabled by Knowledge Graphs, co-located with ISWC},
    eventtitleaddon = {RAGE-KG 2025},
    eventdate    = {2025-11-02/2025-11-06},
    venue        = {Nara, Japan},
    booktitle    = {RAGE-KG 2025: The Second International Workshop on Retrieval-Augmented Generation Enabled by Knowledge Graphs, co-located with ISWC 2025, November 2{\textendash }6, 2025, Nara, Japan},
    publisher    = {{CEUR-WS}},
    issn         = {1613-0073},
    series       = {CEUR Workshop Proceedings},
    language     = {english},
    volume       = {4079}
}

@inproceedings{xie2024adaptive,
    title = "Adaptive Retrieval-Augmented Generation for Conversational Systems",
    author = "Wang, Xi  and
      Sen, Procheta  and
      Li, Ruizhe  and
      Yilmaz, Emine",
    editor = "Chiruzzo, Luis  and
      Ritter, Alan  and
      Wang, Lu",
    booktitle = "Findings of the Association for Computational Linguistics: NAACL 2025",
    month = apr,
    year = "2025",
    address = "Albuquerque, New Mexico",
    publisher = "Association for Computational Linguistics",
    url = "https://aclanthology.org/2025.findings-naacl.30/",
    doi = "10.18653/v1/2025.findings-naacl.30",
    pages = "491--503",
    ISBN = "979-8-89176-195-7"
}

@inproceedings{mallen2023trust,
    title = "When Not to Trust Language Models: Investigating Effectiveness of Parametric and Non-Parametric Memories",
    author = "Mallen, Alex  and
      Asai, Akari  and
      Zhong, Victor  and
      Das, Rajarshi  and
      Khashabi, Daniel  and
      Hajishirzi, Hannaneh",
    editor = "Rogers, Anna  and
      Boyd-Graber, Jordan  and
      Okazaki, Naoaki",
    booktitle = "Proceedings of the 61st Annual Meeting of the Association for Computational Linguistics (Volume 1: Long Papers)",
    month = jul,
    year = "2023",
    address = "Toronto, Canada",
    publisher = "Association for Computational Linguistics",
    url = "https://aclanthology.org/2023.acl-long.546/",
    doi = "10.18653/v1/2023.acl-long.546",
    pages = "9802--9822"
}

@inproceedings{bordes2013translating,
 author = {Bordes, Antoine and Usunier, Nicolas and Garcia-Duran, Alberto and Weston, Jason and Yakhnenko, Oksana},
 booktitle = {Advances in Neural Information Processing Systems},
 editor = {C.J. Burges and L. Bottou and M. Welling and Z. Ghahramani and K.Q. Weinberger},
 pages = {},
 publisher = {Curran Associates, Inc.},
 title = {Translating Embeddings for Modeling Multi-relational Data},
 url = {https://proceedings.neurips.cc/paper_files/paper/2013/file/1cecc7a77928ca8133fa24680a88d2f9-Paper.pdf},
 volume = {26},
 year = {2013}
}

@inproceedings{
kipf2017semi,
title={Semi-Supervised Classification with Graph Convolutional Networks},
author={Thomas N. Kipf and Max Welling},
booktitle={International Conference on Learning Representations},
year={2017},
url={https://openreview.net/forum?id=SJU4ayYgl}
}

@inproceedings{yasunaga2021qa,
    title = "{QA}-{GNN}: Reasoning with Language Models and Knowledge Graphs for Question Answering",
    author = "Yasunaga, Michihiro  and
      Ren, Hongyu  and
      Bosselut, Antoine  and
      Liang, Percy  and
      Leskovec, Jure",
    editor = "Toutanova, Kristina  and
      Rumshisky, Anna  and
      Zettlemoyer, Luke  and
      Hakkani-Tur, Dilek  and
      Beltagy, Iz  and
      Bethard, Steven  and
      Cotterell, Ryan  and
      Chakraborty, Tanmoy  and
      Zhou, Yichao",
    booktitle = "Proceedings of the 2021 Conference of the North American Chapter of the Association for Computational Linguistics: Human Language Technologies",
    month = jun,
    year = "2021",
    address = "Online",
    publisher = "Association for Computational Linguistics",
    url = "https://aclanthology.org/2021.naacl-main.45/",
    doi = "10.18653/v1/2021.naacl-main.45",
    pages = "535--546"
}

@inproceedings{sat2025enhancing,
    title = "Enhancing Large Language Model for Knowledge Graph Completion via Structure-Aware Alignment-Tuning",
    author = "Liu, Yu  and
      Cao, Yanan  and
      Lin, Xixun  and
      Shang, Yanmin  and
      Wang, Shi  and
      Pan, Shirui",
    editor = "Christodoulopoulos, Christos  and
      Chakraborty, Tanmoy  and
      Rose, Carolyn  and
      Peng, Violet",
    booktitle = "Proceedings of the 2025 Conference on Empirical Methods in Natural Language Processing",
    month = nov,
    year = "2025",
    address = "Suzhou, China",
    publisher = "Association for Computational Linguistics",
    url = "https://aclanthology.org/2025.emnlp-main.1061/",
    doi = "10.18653/v1/2025.emnlp-main.1061",
    pages = "20970--20984",
    ISBN = "979-8-89176-332-6"
}

@misc{dhayalkar_attention,
      title={Attention as Binding: A Vector-Symbolic Perspective on Transformer Reasoning}, 
      author={Sahil Rajesh Dhayalkar},
      year={2025},
      eprint={2512.14709},
      archivePrefix={arXiv},
      primaryClass={cs.AI},
      url={https://arxiv.org/abs/2512.14709}, 
}

@article{kiguchi2025multi,
   title={Multi-Modal Integration Analysis of Alzheimer’s Disease Using Large Language Models and Knowledge Graphs},
   volume={13},
   ISSN={2169-3536},
   url={http://dx.doi.org/10.1109/ACCESS.2025.3582853},
   DOI={10.1109/access.2025.3582853},
   journal={IEEE Access},
   publisher={Institute of Electrical and Electronics Engineers (IEEE)},
   author={Kiguchi, Kanan and Tu, Yunhao and Ajito, Katsuhiro and Alnajjar, Fady and Murase, Kazuyuki},
   year={2025},
   pages={113718–113735} }

@misc{zhai2026knowledgefusionbidirectionalinformation,
      title={Knowledge Fusion via Bidirectional Information Aggregation}, 
      author={Songlin Zhai and Guilin Qi and Yue Wang and Yuan Meng},
      year={2026},
      eprint={2507.08704},
      archivePrefix={arXiv},
      primaryClass={cs.CL},
      url={https://arxiv.org/abs/2507.08704}, 
}

@inproceedings{hu2022knowledgeable,
    title = "Knowledgeable Prompt-tuning: Incorporating Knowledge into Prompt Verbalizer for Text Classification",
    author = "Hu, Shengding  and
      Ding, Ning  and
      Wang, Huadong  and
      Liu, Zhiyuan  and
      Wang, Jingang  and
      Li, Juanzi  and
      Wu, Wei  and
      Sun, Maosong",
    editor = "Muresan, Smaranda  and
      Nakov, Preslav  and
      Villavicencio, Aline",
    booktitle = "Proceedings of the 60th Annual Meeting of the Association for Computational Linguistics (Volume 1: Long Papers)",
    month = may,
    year = "2022",
    address = "Dublin, Ireland",
    publisher = "Association for Computational Linguistics",
    url = "https://aclanthology.org/2022.acl-long.158/",
    doi = "10.18653/v1/2022.acl-long.158",
    pages = "2225--2240"
}

@inproceedings{rocktaschel2017end,
 author = {Rockt\"{a}schel, Tim and Riedel, Sebastian},
 booktitle = {Advances in Neural Information Processing Systems},
 editor = {I. Guyon and U. Von Luxburg and S. Bengio and H. Wallach and R. Fergus and S. Vishwanathan and R. Garnett},
 pages = {},
 publisher = {Curran Associates, Inc.},
 title = {End-to-end Differentiable Proving},
 url = {https://proceedings.neurips.cc/paper_files/paper/2017/file/b2ab001909a8a6f04b51920306046ce5-Paper.pdf},
 volume = {30},
 year = {2017}
}

@article{evans2018learning,
author = {Evans, Richard and Grefenstette, Edward},
title = {Learning explanatory rules from noisy data},
year = {2018},
issue_date = {January 2018},
publisher = {AI Access Foundation},
address = {El Segundo, CA, USA},
volume = {61},
number = {1},
issn = {1076-9757},
journal = {J. Artif. Int. Res.},
month = jan,
pages = {1–64},
numpages = {64}
}

@article{van2022analyzing,
   title={Analyzing Differentiable Fuzzy Logic Operators},
   volume={302},
   ISSN={0004-3702},
   url={http://dx.doi.org/10.1016/j.artint.2021.103602},
   DOI={10.1016/j.artint.2021.103602},
   journal={Artificial Intelligence},
   publisher={Elsevier BV},
   author={van Krieken, Emile and Acar, Erman and van Harmelen, Frank},
   year={2022},
   month=jan, pages={103602} }

@inproceedings{manhaeve2018deep,
 author = {Manhaeve, Robin and Dumancic, Sebastijan and Kimmig, Angelika and Demeester, Thomas and De Raedt, Luc},
 booktitle = {Advances in Neural Information Processing Systems},
 editor = {S. Bengio and H. Wallach and H. Larochelle and K. Grauman and N. Cesa-Bianchi and R. Garnett},
 pages = {},
 publisher = {Curran Associates, Inc.},
 title = {DeepProbLog:  Neural Probabilistic Logic Programming},
 url = {https://proceedings.neurips.cc/paper_files/paper/2018/file/dc5d637ed5e62c36ecb73b654b05ba2a-Paper.pdf},
 volume = {31},
 year = {2018}
}

@misc{wang2019satnet,
      title={SATNet: Bridging deep learning and logical reasoning using a differentiable satisfiability solver}, 
      author={Po-Wei Wang and Priya L. Donti and Bryan Wilder and Zico Kolter},
      year={2019},
      eprint={1905.12149},
      archivePrefix={arXiv},
      primaryClass={cs.LG},
      url={https://arxiv.org/abs/1905.12149}, 
}

@inproceedings{giunchiglia2022deep, series={IJCAI-2022},
   title={Deep Learning with Logical Constraints},
   url={http://dx.doi.org/10.24963/ijcai.2022/767},
   DOI={10.24963/ijcai.2022/767},
   booktitle={Proceedings of the Thirty-First International Joint Conference on Artificial Intelligence},
   publisher={International Joint Conferences on Artificial Intelligence Organization},
   author={Giunchiglia, Eleonora and Stoian, Mihaela Catalina and Lukasiewicz, Thomas},
   year={2022},
   month=jul, pages={5478–5485},
   collection={IJCAI-2022} }

@misc{minervini2020differentiable,
      title={Differentiable Reasoning on Large Knowledge Bases and Natural Language}, 
      author={Pasquale Minervini and Matko Bošnjak and Tim Rocktäschel and Sebastian Riedel and Edward Grefenstette},
      year={2019},
      eprint={1912.10824},
      archivePrefix={arXiv},
      primaryClass={cs.LG},
      url={https://arxiv.org/abs/1912.10824}, 
}

@article{elhage2021mathematical,
   title={A Mathematical Framework for Transformer Circuits},
   author={Elhage, Nelson and Nanda, Neel and Olsson, Catherine and Henighan, Tom and Joseph, Nicholas and Mann, Ben and Askell, Amanda and Bai, Yuntao and Chen, Anna and Conerly, Tom and DasSarma, Nova and Drain, Dawn and Ganguli, Deep and Hatfield-Dodds, Zac and Hernandez, Danny and Jones, Andy and Kernion, Jackson and Lovitt, Liane and Ndousse, Kamal and Amodei, Dario and Brown, Tom and Clark, Jack and Kaplan, Jared and McCandlish, Sam and Olah, Chris},
   year={2021},
   journal={Transformer Circuits Thread},
   note={https://transformer-circuits.pub/2021/framework/index.html}
}

@misc{zhong2023mquake,
      title={MQuAKE: Assessing Knowledge Editing in Language Models via Multi-Hop Questions}, 
      author={Zexuan Zhong and Zhengxuan Wu and Christopher D. Manning and Christopher Potts and Danqi Chen},
      year={2024},
      eprint={2305.14795},
      archivePrefix={arXiv},
      primaryClass={cs.CL},
      url={https://arxiv.org/abs/2305.14795}, 
}

@misc{cohen2023evaluating,
      title={Evaluating the Ripple Effects of Knowledge Editing in Language Models}, 
      author={Roi Cohen and Eden Biran and Ori Yoran and Amir Globerson and Mor Geva},
      year={2023},
      eprint={2307.12976},
      archivePrefix={arXiv},
      primaryClass={cs.CL},
      url={https://arxiv.org/abs/2307.12976}, 
}

@misc{wang2024knowledge,
      title={Knowledge Editing for Large Language Models: A Survey}, 
      author={Song Wang and Yaochen Zhu and Haochen Liu and Zaiyi Zheng and Chen Chen and Jundong Li},
      year={2024},
      eprint={2310.16218},
      archivePrefix={arXiv},
      primaryClass={cs.CL},
      url={https://arxiv.org/abs/2310.16218}, 
}

@inproceedings{ruis2020benchmark,
 author = {Ruis, Laura and Andreas, Jacob and Baroni, Marco and Bouchacourt, Diane and Lake, Brenden M},
 booktitle = {Advances in Neural Information Processing Systems},
 editor = {H. Larochelle and M. Ranzato and R. Hadsell and M.F. Balcan and H. Lin},
 pages = {19861--19872},
 publisher = {Curran Associates, Inc.},
 title = {A Benchmark for Systematic Generalization in Grounded Language Understanding},
 url = {https://proceedings.neurips.cc/paper_files/paper/2020/file/e5a90182cc81e12ab5e72d66e0b46fe3-Paper.pdf},
 volume = {33},
 year = {2020}
}

@misc{bahdanau2019systematic,
      title={Systematic Generalization: What Is Required and Can It Be Learned?}, 
      author={Dzmitry Bahdanau and Shikhar Murty and Michael Noukhovitch and Thien Huu Nguyen and Harm de Vries and Aaron Courville},
      year={2019},
      eprint={1811.12889},
      archivePrefix={arXiv},
      primaryClass={cs.CL},
      url={https://arxiv.org/abs/1811.12889}, 
}

@misc{lake2017building,
      title={Building Machines That Learn and Think Like People}, 
      author={Brenden M. Lake and Tomer D. Ullman and Joshua B. Tenenbaum and Samuel J. Gershman},
      year={2016},
      eprint={1604.00289},
      archivePrefix={arXiv},
      primaryClass={cs.AI},
      url={https://arxiv.org/abs/1604.00289}, 
}

@article{smolensky1990tensor,
title = {Tensor product variable binding and the representation of symbolic structures in connectionist systems},
journal = {Artificial Intelligence},
volume = {46},
number = {1},
pages = {159-216},
year = {1990},
issn = {0004-3702},
doi = {https://doi.org/10.1016/0004-3702(90)90007-M},
url = {https://www.sciencedirect.com/science/article/pii/000437029090007M},
author = {Paul Smolensky}
}

@misc{sun2024large,
      title={Large Language Models and Knowledge Graphs: Opportunities and Challenges}, 
      author={Jeff Z. Pan and Simon Razniewski and Jan-Christoph Kalo and Sneha Singhania and Jiaoyan Chen and Stefan Dietze and Hajira Jabeen and Janna Omeliyanenko and Wen Zhang and Matteo Lissandrini and Russa Biswas and Gerard de Melo and Angela Bonifati and Edlira Vakaj and Mauro Dragoni and Damien Graux},
      year={2023},
      eprint={2308.06374},
      archivePrefix={arXiv},
      primaryClass={cs.AI},
      url={https://arxiv.org/abs/2308.06374}, 
}

@article{bricken2023towards,
       title={Towards Monosemanticity: Decomposing Language Models With Dictionary Learning},
       author={Bricken, Trenton and Templeton, Adly and Batson, Joshua and Chen, Brian and Jermyn, Adam and Conerly, Tom and Turner, Nick and Anil, Cem and Denison, Carson and Askell, Amanda and Lasenby, Robert and Wu, Yifan and Kravec, Shauna and Schiefer, Nicholas and Maxwell, Tim and Joseph, Nicholas and Hatfield-Dodds, Zac and Tamkin, Alex and Nguyen, Karina and McLean, Brayden and Burke, Josiah E and Hume, Tristan and Carter, Shan and Henighan, Tom and Olah, Christopher},
       year={2023},
       journal={Transformer Circuits Thread},
       note={https://transformer-circuits.pub/2023/monosemantic-features/index.html}
    }

@misc{zou2023representation,
      title={Representation Engineering: A Top-Down Approach to AI Transparency}, 
      author={Andy Zou and Long Phan and Sarah Chen and James Campbell and Phillip Guo and Richard Ren and Alexander Pan and Xuwang Yin and Mantas Mazeika and Ann-Kathrin Dombrowski and Shashwat Goel and Nathaniel Li and Michael J. Byun and Zifan Wang and Alex Mallen and Steven Basart and Sanmi Koyejo and Dawn Song and Matt Fredrikson and J. Zico Kolter and Dan Hendrycks},
      year={2025},
      eprint={2310.01405},
      archivePrefix={arXiv},
      primaryClass={cs.LG},
      url={https://arxiv.org/abs/2310.01405}, 
}

@misc{park2023linear,
      title={The Linear Representation Hypothesis and the Geometry of Large Language Models}, 
      author={Kiho Park and Yo Joong Choe and Victor Veitch},
      year={2024},
      eprint={2311.03658},
      archivePrefix={arXiv},
      primaryClass={cs.CL},
      url={https://arxiv.org/abs/2311.03658}, 
}

@inproceedings{
wang2024concept,
title={Concept Algebra for (Score-Based) Text-Controlled Generative Models},
author={Zihao Wang and Lin Gui and Jeffrey Negrea and Victor Veitch},
booktitle={Thirty-seventh Conference on Neural Information Processing Systems},
year={2023},
url={https://openreview.net/forum?id=SGlrCuwdsB}
}

@inproceedings{
li2023emergent,
title={Emergent World Representations: Exploring a Sequence Model Trained on a Synthetic Task},
author={Kenneth Li and Aspen K Hopkins and David Bau and Fernanda Vi{\'e}gas and Hanspeter Pfister and Martin Wattenberg},
booktitle={The Eleventh International Conference on Learning Representations },
year={2023},
url={https://openreview.net/forum?id=DeG07_TcZvT}
}

@misc{frady2023variable,
      title={Variable Binding for Sparse Distributed Representations: Theory and Applications}, 
      author={E. Paxon Frady and Denis Kleyko and Friedrich T. Sommer},
      year={2020},
      eprint={2009.06734},
      archivePrefix={arXiv},
      primaryClass={cs.NE},
      url={https://arxiv.org/abs/2009.06734}, 
}

@article{kanerva2009hyperdimensional,
	author = {Pentti Kanerva},
	journal = {Cognitive Computation},
	number = {2},
	pages = {139--159},
	title = {Hyperdimensional Computing: An Introduction to Computing in Distributed Representation with High-Dimensional Random Vectors},
	volume = {1},
	year = {2009}
}

@article{kleyko2022vector,
   title={Vector Symbolic Architectures as a Computing Framework for Emerging Hardware},
   volume={110},
   ISSN={1558-2256},
   url={http://dx.doi.org/10.1109/JPROC.2022.3209104},
   DOI={10.1109/jproc.2022.3209104},
   number={10},
   journal={Proceedings of the IEEE},
   publisher={Institute of Electrical and Electronics Engineers (IEEE)},
   author={Kleyko, Denis and Davies, Mike and Frady, Edward Paxon and Kanerva, Pentti and Kent, Spencer J. and Olshausen, Bruno A. and Osipov, Evgeny and Rabaey, Jan M. and Rachkovskij, Dmitri A. and Rahimi, Abbas and Sommer, Friedrich T.},
   year={2022},
   month=oct, pages={1538–1571} }

\end{document}